*Chapter*

# DEEP NEURAL NETWORKS FOR PATTERN RECOGNITION

## Kyongsik Yun, PhD, Alexander Huyen and Thomas Lu, PhD

Jet Propulsion Laboratory, California Institute of Technology, Pasadena, CA, US

## ABSTRACT

In the field of pattern recognition research, the method of using deep neural networks based on improved computing hardware recently attracted attention because of their superior accuracy compared to conventional methods. Deep neural networks simulate the human visual system and achieve human equivalent accuracy in image classification, object detection, and segmentation. This chapter introduces the basic structure of deep neural networks that simulate human neural networks. Then we identify the operational processes and applications of conditional generative adversarial networks, which are being actively researched based on the bottom-up and top-down mechanisms, the most important functions of the human visual perception process. Finally,

Corresponding Author: kyongsik.yun@jpl.nasa.gov

recent developments in training strategies for effective learning of complex deep neural networks are addressed.

## 1. PATTERN RECOGNITION IN HUMAN VISION

In 1959, Hubel and Wiesel inserted microelectrodes into the primary visual cortex of an anesthetized cat [1]. They project bright and dark patterns on the screen in front of the cat. They found that cells in the visual cortex were driven by a "feature detector" that won the Nobel Prize.

For example, when we recognize a human face, each cell in the primary visual cortex (V1 and V2) handles the simplest features such as lines, curves, and points. The higher-level visual cortex through the ventral object recognition path (V4) handles target components such as eyes, eyebrows and noses. Posterior inferior temporal cortex (pITC) then integrates more complex visual features (faces) with the context (name). Anterior inferior temporal cortex (aITC) more closely matches its name to a particular context (whether this person is female or male, living in California, etc.) [2].

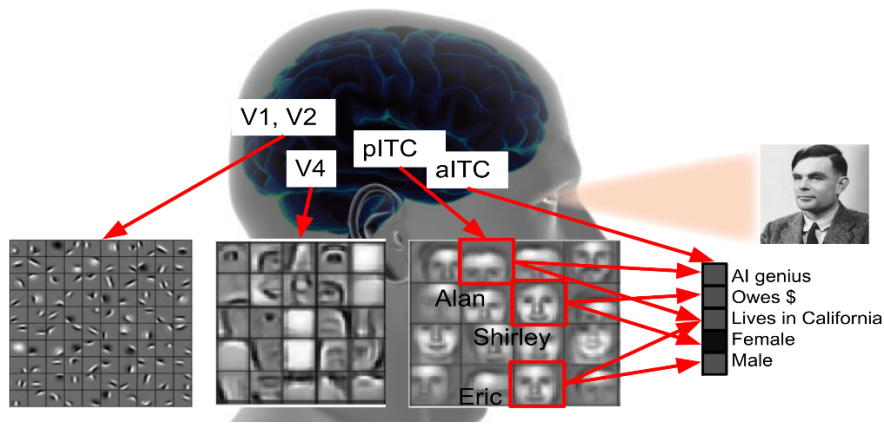

Figure 1. Object recognition in human visual system.

Human visual perception has hierarchical structure as described above and massively parallel processing capability based on thousands of synaptic connections in each neuron. Furthermore, human neural networks feature a winner-take-all framework which selects the most relevant neurons along the spatial dimensions in each layer. These three features (hierarchical structure, parallel processing, winner-take-all framework) inspired to build artificial neural networks, recently further evolved into deep convolutional neural networks. Winner-take-all framework especially adapted to build essential components of deep convolutional networks as max pooling and rectified linear unit (ReLU).

## 2. HUMAN VISION-INSPIRED CONDITIONAL GENERATIVE ADVERSARIAL NETWORKS

Human vision's object recognition can be thought as bottom-up process, which integrates the lower level visual information to make sense out of it [3]. This process can occur in a completely opposite way, top-down process, which is to visualize a new object based on semantic information (imagination). A special type of deep convolutional neural networks, conditional generative adversarial networks can be thought of as an implementation of a human brain process that combines a bottom-up recognition (discriminator network) and a top-down imagination (generator network).

Generative adversarial networks learn the loss of classifying whether the generated output is real or not, while at the same time the networks learn the generative model to minimize this loss [4]. The conditional adversarial networks learn the mapping function from the input to the output as well as learning the loss function to train the mapping [5, 6].

To train the generative adversarial networks, the objective is to solve the min-max game, finding the minimum over $\theta_g$, or the parameters of our generator network G and maximum over $\theta_d$, or the parameters of our discriminator network D.

$$\min_{\theta_g} \max_{\theta_d} \left[ \mathbb{E}_{y \sim p_{data}} \log D_{\theta_d}(y) + \mathbb{E}_{z \sim p(z)} \log \left( 1 - D_{\theta_d} \left( G_{\theta_g}(z) \right) \right) \right] \quad (1)$$

The first term in the objective function (1) is the likelihood, or expectation, of the real data being real from the data distribution $P_{data}$. The log $D(y)$ is the discriminator output for real data $y$. If $y$ is real, $D(y)$ is 1 and log $D(y)$ is 0, which becomes the maximum. The second term is the expectation of $z$ drawn from $P(z)$, meaning the random data input for our generator network (Figure 1). $D(G(z))$ is the output of our discriminator for generated fake data $G(z)$. If $G(z)$ is close to real, $D(G(z))$ is close to 1, and the *log (1-D(G(z)))* becomes very small (minimized).

$$\min_{\theta_g} \max_{\theta_d} \left[ \mathbb{E}_{y \sim p_{data}} \log D_{\theta_d}(y) + \mathbb{E}_{z \sim p(z), x \sim p(x)} \log \left( 1 - D_{\theta_d} \left( G_{\theta_g}(z, x) \right) \right) \right] \quad (2)$$

$$\min_{\theta_g} \max_{\theta_d} \left[ \mathbb{E}_{y \sim p_{data}, x \sim p_{data}} \log D_{\theta_d}(y, x) + \mathbb{E}_{z \sim p(z), x \sim p(x)} \log \left( 1 - D_{\theta_d} \left( G_{\theta_g}(z, x), x \right) \right) \right] \quad (3)$$

$z$ is the random input variable to the generator network. An image can be used as input to the generator network instead of $z$. In the objective function (2), $x$, the real input is a conditional term for the generator. A conditional term $x$ is then added to the discriminator network, as in function (3). The conditional adversarial networks learn the mapping from the input to the output and learn the loss function.

# 3. Precision Multi-Band Infrared Image Segmentation Using Conditional Generative Adversarial Networks

Generative adversarial networks have been applied to the task of segmenting between foreground and background classes in infrared images. One of the key differences between infrared and normal color images is the loss of two channels, with only a single component channel representing the pixel intensity for infrared images. This loss of information results in an increased difficulty of extracting meaningful features during both the training and inference phases. An example is the dynamic characteristics of the foreground and background intensities, which can change based on the amount of infrared radiation captured by the thermal imager. Factors such as sunlight intensity, reflected light and the composition of an object's material contributes to inconsistent features. The intensity of the object and backgrounds in the images vary between frames and wavelengths, which add to the difficulty in achieving a precise segmentation between foreground and background. High noise backgrounds with similar pixel intensities to the object results in ambiguous boundaries that can prove to be difficult for manual segmentation techniques. Low contrast and low intensity features of the objects in these infrared wavelengths also contribute to this difficulty in accurate classification. Small features are areas where precise segmentation is required for accurate pixel classifications, especially in noisy background environments. The nature of generative adversarial networks which feature a generator and discriminator have proven to be a good fit in addressing the difficulties of infrared image segmentation. The relationship of these structures allows generative adversarial networks to capture more relevant features as compared to other network models.

Digital image processing, filtering and thresholding have been used in image segmentation [7, 8]. However, the segmentation accuracy of these methods suffers when applied to images with non-uniform backgrounds. Shallow neural networks have also been used to perform adaptive thresholding [9–11]. A simple neural network with a few hidden layers would still have difficulty processing complex foreground and backgrounds. Traditional neural networks contain one to three hidden layers, whereas deep learning models can contain as many as 100 hidden layers. Typically, DNN models learn classification tasks from large labeled datasets of images, texts, or audio through a process called feature extraction. The multi-layered nature of a DNN allows for each layer to encode distinctive features. When classifying features in an image, some layers can extract edge features while others can extract texture. As the layers advance towards the output layer, encoded features become more complex.

One of the drawbacks of DNNs is the requirement of a large number of training samples. It is time consuming to prepare training samples and ground truth data. Sometimes it is impossible to obtain many training samples. The DNN model discussed here makes use of conditional generative adversarial networks that learns a mapping from an input image to an output image [12]. By the nature of generative adversarial networks, the model generates a loss function directly from the training data. Conditional generative adversarial networks are conditioned to learn mappings from a dataset of paired images, a real image and its ground truth, which is the precise segmentation boundary of the object. The model is conditioned to take an infrared image of an object as an input and produce a binary mask of the object as an output. Using the feedback from the discriminator, the generator is able to produce images that resemble the desired ground truths from the input image.

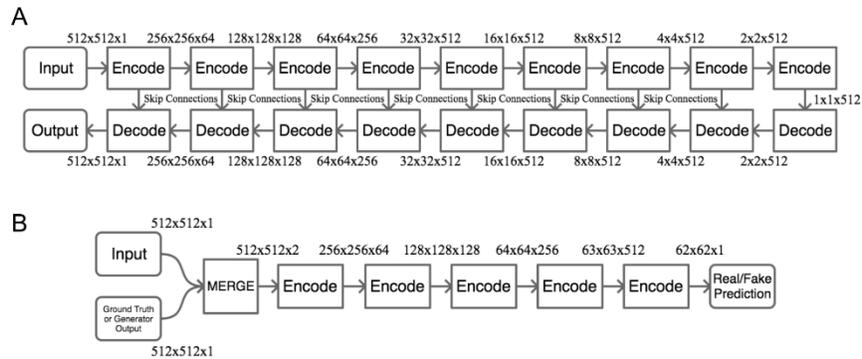

Figure 2. Conditional generative adversarial networks for object segmentation. (A) the U-Net generator to produce output images based on input images maintaining structural similarity between the input and output images by using skip connections; (B) the discriminator to determine whether the generator output is real or fake.

Figure 2 shows the network architecture of the U-Net [12], an encoder/decoder network with skip connections (Figure A) and the discriminator (Figure B). The generator architecture consists of a nine-layer encoder and nine-layer decoder with skip connections between each encoder and decoder layer. An additional layer was added to both the encoder and decoder to increase the resolution from 256x256 to 512x512. As a result, a skip connection was added between the first encoder layer and the eighth decoder layer.

An input image is resized into a 512x512 matrix to match the size of the input layer. The input image comes in the first layer in the top left. The data flows through the network and the segmented map comes out of the top right side of the generator. In each encoder layer, the network performs a 4x4 convolution, batch normalization, and an activation function (leaky rectified linear unit (LeakyReLU)). The network performs a 4x4 deconvolution, batch normalization, and an activation function (rectified linear unit (ReLU)) for each decoder layer.

Artificial training images are used to enhance the segmentation accuracy of misclassified features of certain objects. The target is removed and the surrounding area's texture is taken to fill in the missing target creating an image with only the desired feature. After removing the object, the cropped feature is overlaid on top of the background image. Enhancement images are designed to increase the loss contribution of specific object features during the model's retraining. By increasing the number of training images containing these emphasized loss contributions, the desired features are enhanced in the DNN.

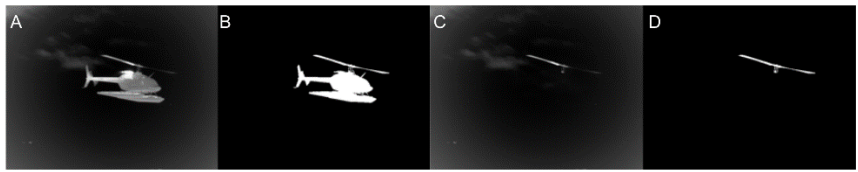

Figure 3. Example of an Augmented FF Dataset, (A) Original image, (b) GT, and an example of an Augmented feature cropped Dataset: (c) missing rotor part, (d) GT of the missing rotor.

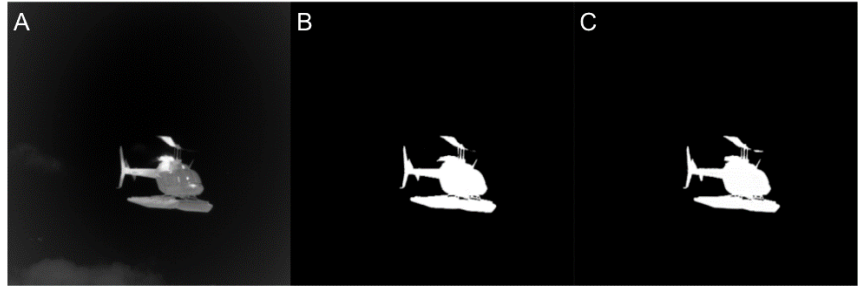

Figure 4. Retraining a DNN to enhance features: (a) the original image; (b) initial DNN output missing the right side of rotor blade; (c) the retrained DNN output showing the missing rotor blade part is reconstructed.

GraphCut is a semi-manual segmentation technique which divides the entire frame into regions and weighs each based on the similarity to nearby regions. Edges where the weights are low are separated into foreground and background. A manual selection of samples in the foreground and background are needed by GraphCut to distinguish the class of its regions. More manual sample selections can be added to further refine the segmentation, but if the similarity between some regions of the frame is high, no matter how many manual sample selections are increased, false classification will occur.

Binary segmentation masks can be created by applying an automatically generated threshold. Iterative inter-means or IsoData [8] was used for the automatic threshold to perform the segmentation. This algorithm separates foreground and background with a starting threshold, which is then updated by calculating an average of two averages from values of the pixels above and below the initial threshold. The initial threshold is updated until it is higher than the average of two averages from pixels higher and lower than it. ImageJ [13] was used to perform segmentation by manually thresholding the image. The threshold value was selected based on the accuracy of segmented features and amount of noise generated. A higher threshold tends to remove a majority of noise in the segmentation but misses low intensity features like the antennae and rotor blades.

A small scale neural network was used to perform infrared image segmentations [10]. This neural network consisted of one input layer, one hidden layer and one output layer. The hidden layer of this simple neural network contained only 25 neurons. The simple neural network gives decent segmentation results (9.2% XOR error) compared to the previous GraphCut (19.0% XOR error) and thresholding methods (10.3% XOR error). A drawback of this small scale neural network is the lack of generalization across intensity changes in the foreground and background.

DeepLab v2 (ResNet-101) [14] is an implementation of deep convolutional networks for the purpose of semantic segmentation. The specific implementation used here features atrous convolution and atrous spatial pyramid pooling, without the application of conditional random fields for post processing. Atrous convolution refers to retrieving full resolution features by using upsampled filters in the convolutional layers. Atrous spatial pyramid pooling (ASPP) is used during the feature abstraction stages to increase resilience against shifts in scale by utilizing multiple filters with various sampling rates [14]. Figure 5 showed the results of the segmentation by Deeplab in Figure 5 and the modified conditional generative adversarial networks (JPL's segmentation) in Figure 5. Deeplab was capable of identifying the object and separating the object from background. However, the Deeplab model was not able to draw an accurate boundary of the object, missing small details of the object.

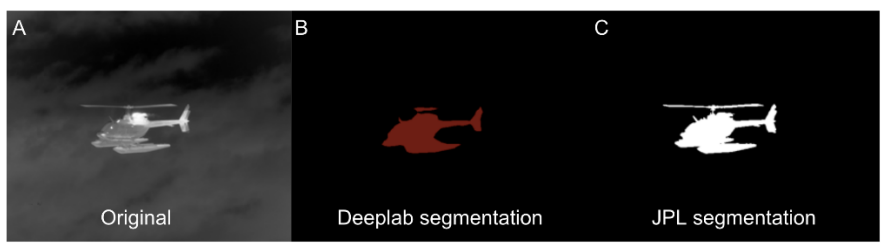

Figure 5. Segmentation results: (A) Original image; (B) Deeplab ResNet segmentation output; (C) JPL segmentation output using conditional generative adversarial networks.

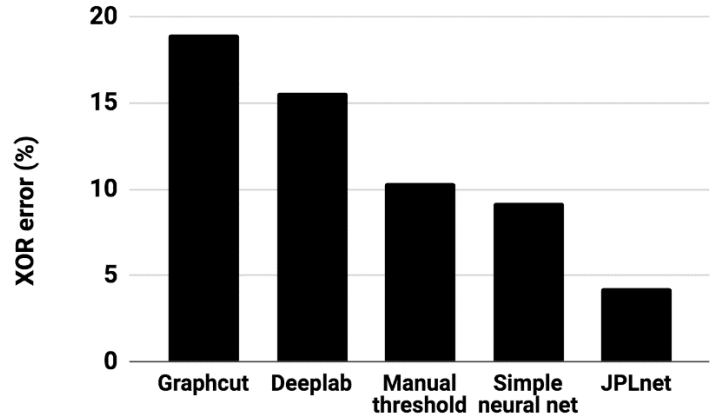

Figure 6. Comparison of the segmentation results of Graphcut, Deeplab-ResNet, manual thresholding, simple neural network and JPL network (modified conditional generative adversarial networks).

The modified conditional generative adversarial networks model has shown to be able to generate significantly more accurate segmentation masks than the thresholding methods, GraphCut, DeepLab v2 model, and the simple neural network model. Although the simple neural network achieved an average XOR percentage of 9.2% with less than a minute of training, manual input is required. Manual selection of coordinates to create a bounding box around the object is needed for both training and inference. The simple neural network also has difficulty with segmenting fine features around the low intensity regions.

The conditional generative adversarial networks architecture has demonstrated the ability to precisely segment objects from a noisy background in infrared wavelength images. Low intensity features and gradient edges are clearly extracted by the model where methods like iterative inter-means thresholding, manual thresholding, and GraphCut generally fail. The DNN can be trained by a relatively small number of training images due to the conditional generative adversarial networks architecture. It demonstrated robust object detection, identification and precision segmentation in highly cluttered backgrounds. Other DNN architectures focused on semantic segmentation are able to generate segmentation masks better than thresholding methods but often miss major features of the object.

## 4. Occluded Object Reconstruction Using Conditional Generative Adversarial Networks

Conditional generative adversarial networks can be used to reconstruct partially occluded objects [15]. Occluded object reconstruction can be useful to provide situational awareness information especially to next-generation first responders to help them perform safe, healthy, and successful missions. More than 30,000 firefighters are injured each year during firefighting operations [16, 17]. Slips, trips, and falls cause a large number of firefighter injuries, reinforcing the need for advanced ground support for firefighters [18, 19].

Occluded object reconstruction can provide the "see-through" feature in augmented reality glasses [20]. If firefighters can see through the fire, it will help them to navigate the incident more safely and efficiently. Moreover, firefighters can quickly find victims partially occluded by objects. Estimating the size and shape of a fire in a partially occluded situation is particularly important in that it can predict a sudden fire explosion (flashover phenomenon) [21]. If firefighters are provided with see-through capability, they can be safe from dangerous environments and fully understand the surrounding situation.

In this study, the authors used generative adversarial networks to train associations between various images of flammable and hazardous objects and their partially occluded counterparts. Figure 7 show the overall architecture, including input occluded images, generated output reconstruction images, and ground truth images. While the original generative adversarial networks generate real-looking images from the random noise input $z$, the conditional generative adversarial networks can find the association of the two images (i.e., original and occluded), and transform or reconstruct the real input image into another image.

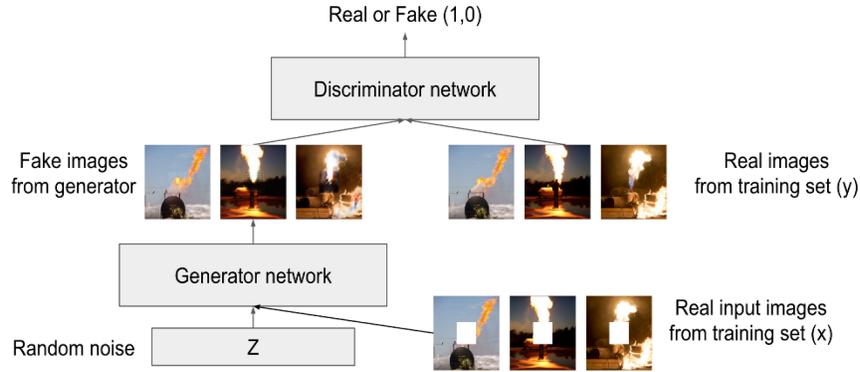

Figure 7. Conditional generative adversarial networks architecture. The generator network uses the occluded real image as input to output a reconstructed fake image. The generated image is compared with the real image in the discriminator network that outputs the classification of the real or the fake. With adversarial training, the generator produces a more realistic image, and the discriminator more accurately distinguishes between real and fake.

The goal was to reconstruct partially occluded objects in the image, so it can be helpful to train the basic structure of the actual objects (fire, gas tank, roof, etc.) and to maintain the structural similarity of the input and output images. Therefore, the authors used U-Net architecture [22] for our generator based on an encoder-decoder network that was gradually downsampled and upsampled for efficiency. The network then applied the skip layer. That is, each downsampling layer is sent to and connected to the corresponding upsampling layer. Finally, the upsampling layer could directly learn important structural features from the downsampling layer.

The authors collected 100 images from Google Images (keywords: gas tank fire) (Figure 8). They then added artificial occlusion to the image and used it as input. We used 80 image pairs randomly selected for training and 20 image pairs for testing.

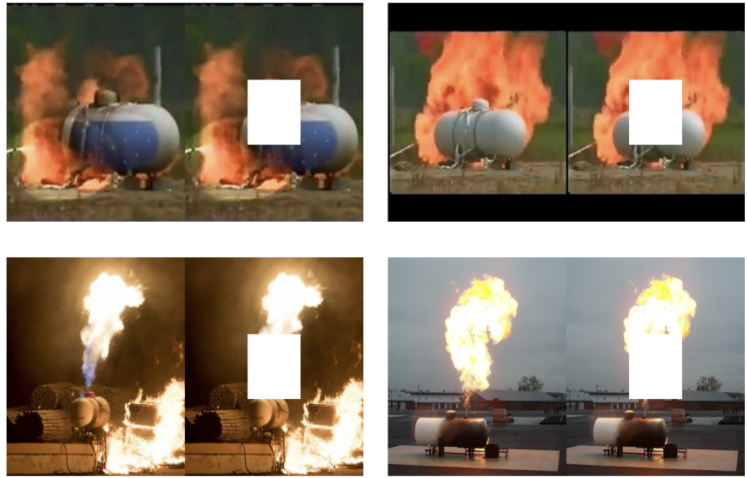

Figure 8. Representative training image pairs for our conditional generative adversarial networks. Propane gas tank images in various situations were used for training. The original image was entered into the discriminator network and the occluded copy was used as the generator network input.

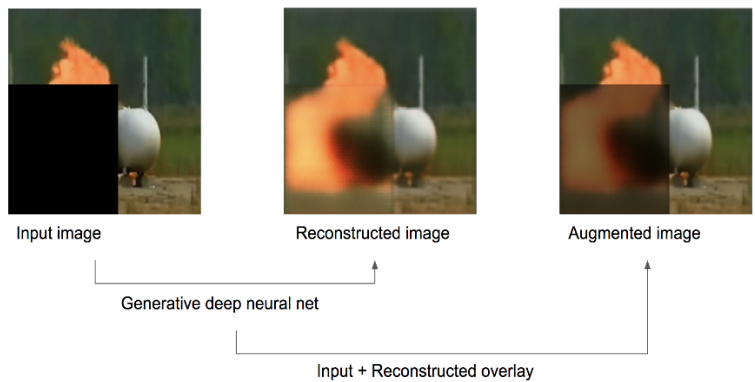

Figure 9. Reconstructing a partially occluded object for first responders using conditional generative adversarial networks and augmented reality glasses. The reconstructed image was overlaid on the input image and used as a visual enhancement feature for the first responder.

After training conditional generative adversarial networks, the system reconstructed test images of partially occluded flammable objects. Finally, the reconstructed image was superimposed on the input image to provide "transparency" (Figure 9). Representative reconstructions are shown in

Figure 10. Five out of 20 test images were incorrectly reconstructed as shown in Figure 10C (25% error rate).

Conditional generative adversarial networks have shown good results on small data sets (fewer than 100 samples) [5, 15, 21]. The advantage of a small datasets is the speed of training. The findings of the study were based on less than five hours of training on a single NVIDIA GTX1080 GPU. Because the time it takes to convert the test output image is shorter than 200 milliseconds, this study can be applied as a real solution that incorporates improved insight into the augmented reality glasses of the next generation first responder.

Previous studies have attempted to reconstruct an occluded object in augmented reality by solving mathematical models [20, 23]. Using deep learning-based conditional generative adversarial networks, we could reconstruct complex backgrounds and multiple objects more easily than mathematical models. Deep learning can be thought of as a nonlinear numerical model. Mathematical models may not be accurate in a complex real world.

If the objective is to provide a warning and help identify the victim in a fire situation, one may argue that we can provide the necessary specific information directly, without having to reconstruct the partially occluded object as in this study. It may be true, but by providing a reconstruction of the occluded object, first responders can discover other significant risks that cannot be identified by a warning algorithm. Moreover, in terms of explanatory and transparent solutions, lower level perceptual information (visualization of occluded parts) is more intuitive and faster than warning signals [24–26].

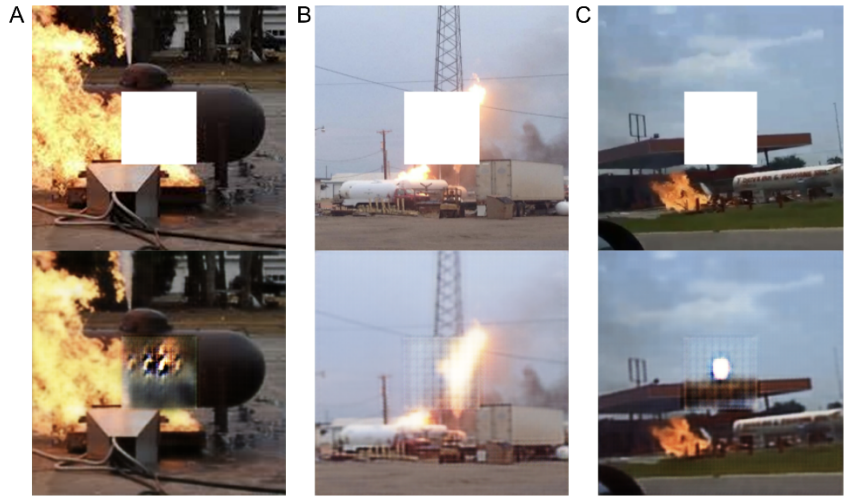

Figure 10. Representative correct and incorrect reconstruction examples. (A) The fire pattern was not generated correctly. (B) The fire pattern was accurately generated. The transmission tower located in the background behind the fire was not reconstructed. (C) A false fire pattern was created (negative example). The roof was generated correctly.

The reason for not reconstructing the objects correctly in some test images (Figure 4C) may be due to a lack of variability in the training data set. As the next step, the number and variability of training images should be increased to deal with various types of gas tanks and fire events. Contextual information, including building structure, fire cause, and gas/chemical sensor information, will help provide more accurate semantic reconstruction. Image changes over time and perspective changes can also be added to increase accuracy.

## 5. Image Enhancement from Visual to Infrared Using Conditional Generative Adversarial Networks

Conditional generative adversarial networks can be used to enhance images. Deep learning based image enhancement can be applied to provide contextual insights for firefighters to predict flashover phenomena [27–29]. Flashover is an example of a fire that spreads very quickly across a gap because of the intense heat. Flashover is the most frightening phenomenon among firefighters. In particular, firefighters need years of training and experience to identify, predict, and plan how to engage flashovers [30, 31].

Previous studies constructed mathematical models of flashover based on variables, including fuel, temperature, and ventilation [32–34]. A previous study applied a numerical model using oxygen concentration and mean flame surface size to predict flashover time, and the authors predicted flashover within 70 seconds before it occurred [35]. Another study developed a method to estimate the temperature of a fire before flashover [36]. However, these computational models of flashover fire require structural information of the area and temperature/oxygen data using special sensors that are not practical for real fire mission applications.

The index of flashovers includes dark smoke, high heat and rollover (angel fingers) and can be quantified by color, size and shape. Firefighters use hand-held thermal imaging cameras to detect and analyze the quality of fire in dark situations [37]. Hand-held cameras require one hand for location and operation, leaving only one hand for other tasks. Due to the lack of a properly used thermal imaging camera, it has contributed to the injury and death of a firefighter [18, 38]. Moreover, due to the budget constraints of the fire department, the infrared camera system is not currently available for all firefighters [39].

To solve this problem, we used a standard body camera and analyzed the color video stream. We applied generative adversarial neural networks [4] to enhance very dark fire and smoke patterns. We then monitored the dynamic changes of smoke and fire. Finally, we provided information on when flashover could occur.

The basic idea comes from the fact that experienced firefighters can identify fire and smoke quality visually even in very dark conditions [40–42]. They can imagine a thermal image through a regular color image. So we applied image-to-image conversion techniques using conditional adversarial networks [5] that have been shown to be effective in synthesizing photos from label maps, reconstructing objects from edge maps, and colorizing images.

Conditional adversarial network training requires a thermal image (input) that matches a regular color image (output). The network maps the input image to the output image. Flashover training videos were provided by Lewisville Fire Department and Texas State Fire Marshal's Office. These videos show the fire generation and evolution through flashover at an early stage using the burn pod made by the department. The video includes both a regular video camera and a thermal camera view that can be seen by firefighters. A total of 40 frames of regular and thermal image pairs were used in this preliminary study (30 images for training, 10 images for the test).

The authors calculated the number of pixels in the red, green, and blue channels. The yellow channel is defined as the average of the red and green channels. Red is 300-500°F, yellow is 200-300°F, green is 100-200°F, and blue is 0-100°F depending on the temperature representation of the infrared camera. This drew the time variation of the number of pixels in each channel.

In the test of the generative network, ten images were used as inputs from the beginning of the fire to the flashover transition (0-200 seconds). The output image was then created (Figure 11). During the early stages of the fire, the general structure of the objects (tables and sofas) and mild smoke were identified (Figure 11A). Even in very dark environments, the network successfully generated hot smoke from the input image (Figure 11B). In a flashover situation, a high heat was calculated and the table object was still identified (Figure 11C).

The output images were further analyzed by counting pixels of each thermal component (red, yellow, green, blue) and then their temporal changes were plotted (Figure 12). Based on the temporal changes of high temperature components (300-500°F and 200-300°F), the authors predicted a flashover as early as 55 seconds before it occurred.

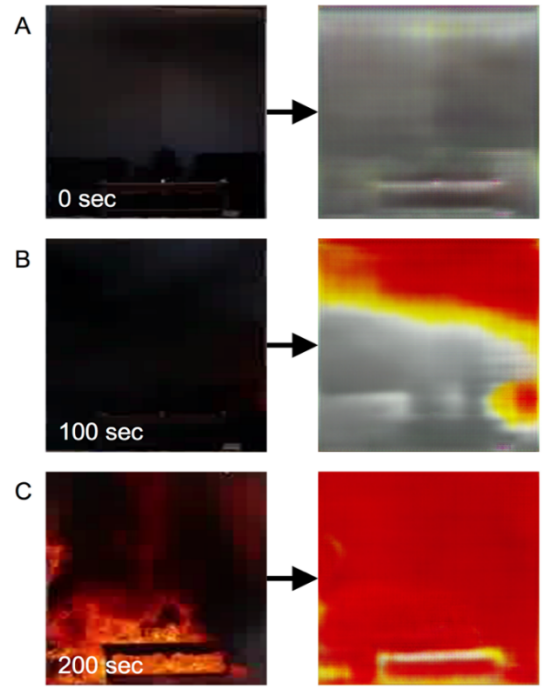

Figure 11. Representative images of transformation from the visual to the thermal images using generative adversarial networks.

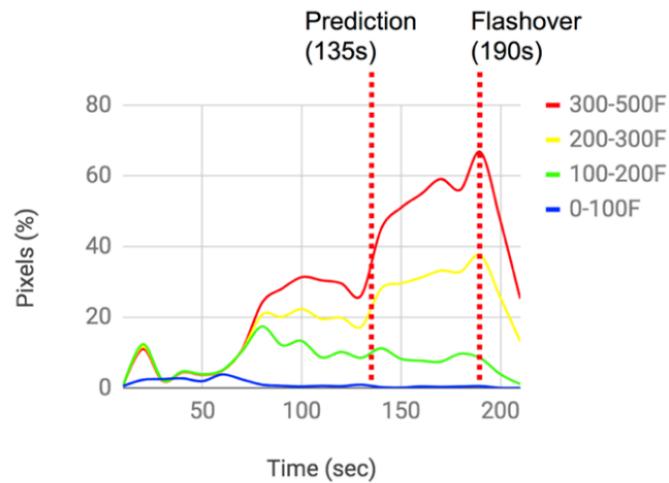

Figure 12. Size of smoke and flame over time. Based on the rate of changes in smoke and flame size, flashover fire was predicted as early as 55 seconds before it occurred.

In this study, the authors showed that regular body camera images can be transformed and enhanced to represent thermal information, and this technique can be used to predict flashover 55 seconds before it occurs. In order to fully understand the process of image enhancement from the regular visual image to the thermal image, we visualized the transformed intermediate results in each latent space of the networks (Figure 13). In the early part of the space, representative features (i.e., fire and smoke) were improved. In the deeper part of the space, fire and smoke were further subdivided by the estimated temperature. Then representative objects (i.e., tables and sofas) were detected. As a result, thermal information and representative objects of fire and smoke were reconstructed in the final output image. Visualization of the hidden layers of neural networks is particularly useful for understanding the internal dynamics of the whole process, which has often been considered a black box [43, 44].

A previous study presented a thermal camera system integrated into a firefighter helmet equipped with an OLED display [45]. The authors' proposed algorithm can also be applied to predict flashover by automatically analyzing thermal images and quantifying the time changes of thermal content of smoke and fire. As for information overload, a high level of information (i.e., flashover timing prediction) can be provided to firefighters without overlaying the heat/occluded information on the display. This is best not to disturb the firefighter's view, but still provides the most important information.

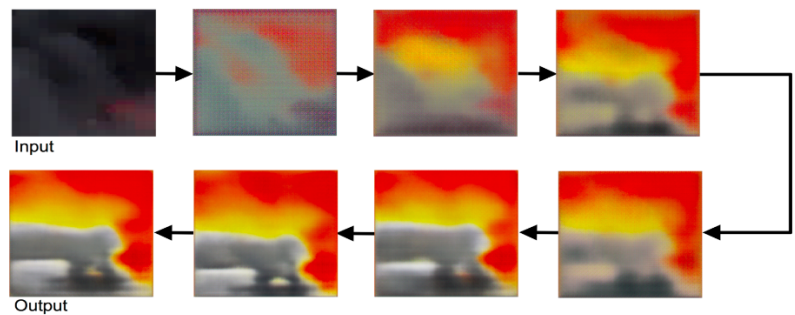

Figure 13. Representative latent-space visualizations of the generative adversarial networks for image enhancement.

# 6. Data Augmentation for Training Deep Neural Networks

One of the major challenges in deep learning is retrieving sufficiently large labeled training datasets, which can become expensive and time consuming to collect. A unique approach to training segmentation is to use Deep Neural Network (DNN) models with a minimal amount of initial labeled training samples. The procedure involves creating synthetic data and using image registration to calculate affine transformations to apply to the synthetic data. The method takes a small dataset and generates a high-quality augmented reality synthetic dataset with strong variance while maintaining consistency with real cases. Results illustrate segmentation improvements in various target features and increased average target confidence. The motivation for synthetic data is due to lack of available labeled data, which is recommended in large volumes for supervised deep learning. Particularly for resource limited research groups, data is difficult to collect due to monetary and time expenses. Furthermore, deep learning is susceptible to poor training efficiency, and fully training a deep neural network requires a significant amount of time especially if computing resources are limited [46]. The rate of progress in this field can be improved by making further progress in data accessibility and training optimization.

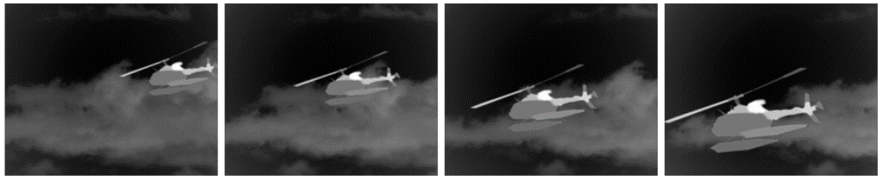

Figure 14. Example of an augmented reality synthetic video sequence created through successive transformations and superimposing the resulting images onto a sequence of background frames.

Data augmentation is a method for increasing the size of a dataset and the approach has demonstrated performance benefits in deep learning even with synthetic data [47]. Data augmentation has been proven to improve image segmentation models and this is especially beneficial when working with small training datasets [48, 49]. A dataset can be augmented using a variety of processing tools, mainly by adding noise to an image, performing geometric transformations, or adjusting the original colors and contrast. As a result, the augmented dataset artificially introduces new information that can improve generalization of the network and prevent overfitting. In situations where training data is expensive to obtain and manually label, synthetic data can provide valuable additional training sets. For segmentation tasks, synthetic targets are created from a template and multiplied through the application of transformations. The resulting synthetic training targets can be easily binarized through a global threshold to create their corresponding labeled masks. This process is both efficient and inexpensive in generating training samples and their labels. The quality and size of the dataset are arguably just as important as the optimization of network parameters [50].

External videos that contain an object demonstrating a desired motion were used as a reference to retrieve the transformation matrices to apply to the synthetic data. These videos do not necessarily have to be of the same style representing the model; only the dynamic properties of the object need to be representative. Once the synthetic data has been transformed, the sequence of frames can be superimposed onto a sequence of background images as shown in Figure 15.

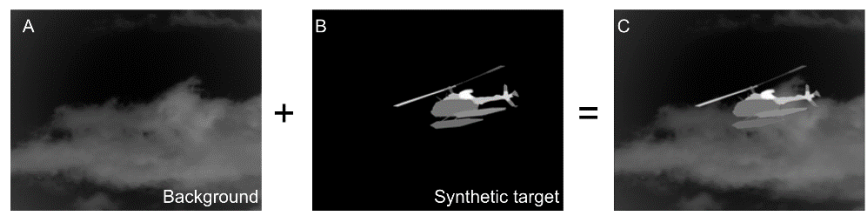

Figure 15. Augmentation process: (a) Preprocessed background frame, (b) Synthetic target frame, (c) Superposition of Target and background frames.

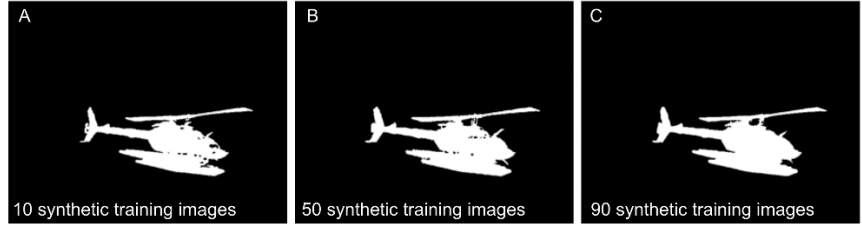

Figure 16. Example of progression of segmentations as more images are included: (a) Result from initial training set (10 image training set), (b) Result from second retraining iteration (50 image training set), (c) Result from fourth retraining iteration (90 image training set).

**Table 1. Synthetic data training average test accuracy results on the 116-image test set and the 154-image test set**

| Number of synthetic training images | 10 | 30 | 50 | 70 | 90 | 110 |
|---|---|---|---|---|---|---|
| Segmentation test accuracy | 89.30% | 90.86% | 90.84% | 91.17% | 91.99% | 92.21% |

Figure 16 shows the progression of improved accuracy as the synthetic training samples are added to the training set. Consistent improvements are expected as images are added. More training images are added to the training data set based on the previous training's performance, but it is difficult to verify whether the new images will be beneficial to the specific problem set without testing after propagating them through the CNN.

The training set includes an initial set of synthetic target frames. This set of data is transformed using the proposed method to augment the training dataset. Table 1 shows the results of test accuracy vs number of synthetic images added to the training set. The DNN performs better when more augmented synthetic images are used to train the DNN.

AR is a useful method for creating data for DNN training as it can automatically guide data augmentation by tracing the real-world dynamics of an object in a reference video. State-of-art AR has also allowed for a variety of applications in simulation, education and medical [51]. The results from training a DNN using AR for dataset augmentation has demonstrated effectiveness in precision segmentation of infrared images. As more images were added to the datasets, network performance increased and further supports the concept of using data augmentation to improve the quality of training datasets. For single class, low-resolution infrared images, our AR approach score achieved an accuracy of 92.21% when using synthetic targets and 92.78% when using real targets. The 110 image synthetic training dataset performed within 1% similar accuracy as the 110 image real dataset, which also validates the effectiveness in training a DNN with synthetic data. The results of this suggest that AR can be applied as a convenient tool for training a DNN for precision infrared image segmentation.

## 7. Incremental Training

Incremental training provides a practical solution for deep neural networks where training from scratch is time consuming and resource intensive. Convolutional neural networks (CNN) have proven to be able to produce world class results in multiple domains such as computer vision, natural language processing, and speech recognition. CNNs leverage large amounts of training data to expand its ability to generalize and adapt to its current problem set. Compared to previous architectures that required many experts to hand-engineer features, CNNs can process raw data and generate relevant features autonomously. These features are automatically extracted from the training data by the CNN, with only the architecture of the neural network being pre-defined. Despite the effectiveness of these neural networks in adapting to new problems, some drawbacks are present. Deep neural networks are computationally demanding, take large amounts of training data and requires lots of power and time to fully train. Hyper parameters such as learning rates, exponential decay rates, and loss function types are difficult to select and calibrate. Hardware limitations forces training data sets to be split into batches instead of propagating through the network all at once. This lengthy process is unacceptable to repeat from scratch every time the network needs to be trained with new data. Incremental training can potentially maximize the accuracy of the CNN by addressing these issues. One application of incremental training can be applied to selectively fine-tune certain missed features without requiring training with the entire data set. Instead of adding in new data and relying on random batching to update the new features, incremental training can be used to focus on what was missed in previous batches. This method of incremental training is optimal during the fine-tuning stages of training with a relatively low amount of training samples, while the benefits of training from scratch increases as the number of training data remains relatively unchanged and exceptionally large. Various approaches utilizing incremental training for support vector machines and hidden Markov models are found in [52–55].

In an experiment with the results plotted in Figure 17 the conditional generative adversarial networks model was trained using an initial subset of 12 images from 68 ground truths. This initial training on the 12 images was trained in 1.73 hours to 1000 epochs. A second training session was done by adding 8 new images to the previously used 12-image set from the 68-image set. The epoch with the lowest L1 validation score from the 12-image training session was selected as the weight matrix to begin retraining for the 20-image training. This 20-image training session was run for 1.96 hours to 1000 epochs. The same process was repeated for the final training, with 16 new images being added to the 20-image dataset. The lowest L1 validation scoring weight matrix from the 20-image set was used to continue training for this new 36-image set. This 36-image training was done for 2.9 hours to 1000 epochs. A final training was done for another 4.05 hours to 1000 epochs on top of the weight matrix from the 36-image set with the lowest L1 validation score using the entire set of 68 images. This process of increasing the training set in batches and retraining using a previous optimal weight matrix generated more accurate results than the regular method of training. In comparison, using the conventional method of training from scratch with the entire set of 68 images to 10,000 epochs for 42.45 hours achieved an average XOR error percentage of 4.98% compared to 10.64 hours of training for a total of 4000 epochs and a resulting average XOR error percentage of 4.23%. This experiment shows the effectiveness of incremental training with new training images, compared to training with an entire dataset with much longer running times.

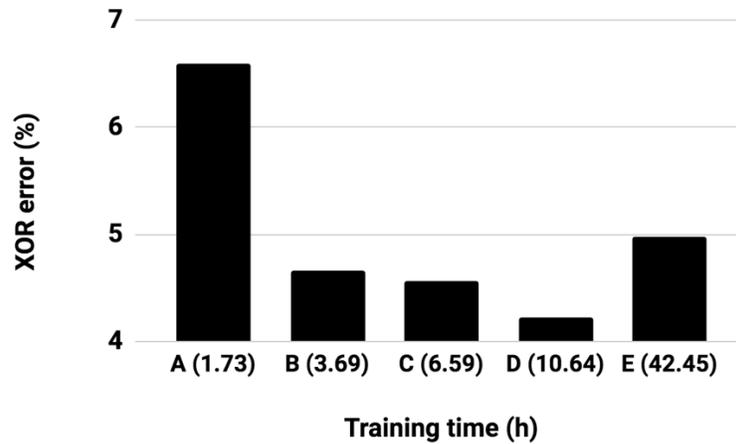

Figure 17. Performance (error) comparison of retraining process: Four-step retraining results (A, B, C, D) compared against single training results (E). The 4-step training achieved lower error rate in a quarter training time as compared to a single training session. The XOR error percentages for each training iteration was generated by calculating the average results from testing on an 11-image standard test dataset that was not included in the training dataset.

## CONCLUSION

Deep convolutional neural networks, especially conditional generative adversarial networks tremendously improved the accuracy in image classification and enhancement, and object detection and segmentation. The system imitates human learning about the laws of physics through experience by learning the shape and characteristics of objects for precise segmentation and enhancement. As human learning, incremental learning is effective in training deep convolutional neural networks. This system will provide important support to enable humans to perform any task more accurately and efficiently.

## ACKNOWLEDGMENT

The research was carried out at the Jet Propulsion Laboratory, California Institute of Technology, under a contract with the National Aeronautics and Space Administration. The research was funded by the U.S. Department of Homeland Security Science and Technology Directorate Next Generation First Responders Apex Program (DHS S&T NGFR) under NASA prime contract NAS7-03001, Task Plan Number 82-106095, and by the U.S. Department of Defense, Test Resource Management Center, Test & Evaluation/Science and Technology (T&E/S&T) Program under NASA prime contract NAS7-03001, Task Plan Number 81-12346.

## REFERENCES

[1]    Hubel, D. H. and Wiesel, T. N. "Shape and arrangement of columns in cat's striate cortex," *The Journal of physiology*, vol. 165, no. 3, pp. 559–568, 1963.

[2]    Vandenberghe, R., Price, C., Wise, R., Josephs, O., and Frackowiak, R. S. J. "Functional anatomy of a common semantic system for words and pictures," *Nature*, vol. 383, no. 6597, p. 254, 1996.

[3]    Yun, K. and Stoica, A. "Improved target recognition response using collaborative brain-computer interfaces," presented at the Systems, Man, and Cybernetics (SMC), 2016 IEEE International Conference on, 2016, pp. 002220–002223.

[4]    Goodfellow, I. et al., "Generative adversarial nets," in *Advances in neural information processing systems*, 2014, pp. 2672–2680.

[5]    Isola, P., Zhu, J. Y., Zhou, T., and Efros, A. A. "Image-to-image translation with conditional adversarial networks," *arXiv preprint*, 2017.

[6]    Mirza, M. and Osindero, S. "Conditional generative adversarial nets," *arXiv preprint arXiv:1411.1784*, 2014.

[7]   Boykov, Y., Veksler, O., and Zabih, R. "Fast approximate energy minimization via graph cuts," *IEEE Transactions on pattern analysis and machine intelligence*, vol. 23, no. 11, pp. 1222–1239, 2001.

[8]   Ridler, T. W. and Calvard, S. "Picture thresholding using an iterative selection method," *IEEE trans syst Man Cybern*, vol. 8, no. 8, pp. 630–632, 1978.

[9]   Lu, T. et al., "Cross-correlation and image alignment for multi-band IR sensors," in *Optical Pattern Recognition XXVII*, 2016, vol. 9845, p. 984505.

[10]  Lu, T. et al., "Intelligent multi-spectral IR image segmentation," in *Pattern Recognition and Tracking XXVIII*, 2017, vol. 10203, p. 1020303.

[11]  Lu, T. and Mintzer, D. "Hybrid neural networks for nonlinear pattern recognition," *Optical Pattern Recognition*, pp. 40–63, 1998.

[12]  Isola, P., Zhu, J. Y., Zhou, T., and Efros, A. A. "Image-to-image translation with conditional adversarial networks," *arXiv preprint*, 2017.

[13]  Rasband, W. S. "ImageJ: Image processing and analysis in Java," *Astrophysics Source Code Library*, 2012.

[14]  Chen, L. C., Papandreou, G., Kokkinos, I., Murphy, K., and Yuille, A. L. "Deeplab: Semantic image segmentation with deep convolutional nets, atrous convolution, and fully connected crfs," *IEEE transactions on pattern analysis and machine intelligence*, vol. 40, no. 4, pp. 834–848, 2018.

[15]  Yun, K., Lu, T., and Chow, E. "Occluded object reconstruction for first responders with augmented reality glasses using conditional generative adversarial networks," in *Pattern Recognition and Tracking XXIX*, 2018, vol. 10649, p. 106490T.

[16]  Bowyer, M. E., Miles, V., Baldwin, T. N., and Hales, T. R. *"Preventing deaths and injuries of fire fighters during training exercises,"* 2016.

[17]  Butler, C., Marsh, S., Domitrovich, J. W., and Helmkamp, J. "Wildland firefighter deaths in the United States: A comparison of

existing surveillance systems," *Journal of occupational and environmental hygiene*, vol. 14, no. 4, pp. 258–270, 2017.

[18] Karter, M. J. and Molis, J. L. *US Firefighter Injuries-2012. National Fire Protection Association, Fire Analysis and Research Division, Quincy, MA*. 2012.

[19] Smith, D. L. *Firefighter fatalities and injuries: the role of heat stress and PPE*. Firefighter Life Safety Research Center, Illinois Fire Service Institute, University of Illinois, 2008.

[20] Azuma, R., Baillot, Y., Behringer, R., Feiner, S., Julier, S., and MacIntyre, B. "Recent advances in augmented reality," *IEEE computer graphics and applications*, vol. 21, no. 6, pp. 34–47, 2001.

[21] Yun, K., Bustos, J. and Lu, T. "Predicting Rapid Fire Growth (Flashover) Using Conditional Generative Adversarial Networks," *Electronic Imaging*, vol. 2018, no. 9, pp. 1–4, 2018.

[22] Ronneberger, O., Fischer, P., and Brox, T. "U-net: Convolutional networks for biomedical image segmentation," in *International Conference on Medical image computing and computer-assisted intervention*, 2015, pp. 234–241.

[23] Lee, J., Hirota, G., and State, A. "Modeling real objects using video see-through augmented reality," *Presence: Teleoperators & Virtual Environments*, vol. 11, no. 2, pp. 144–157, 2002.

[24] Gunning, D. "Explainable artificial intelligence (xai)," *Defense Advanced Research Projects Agency (DARPA), nd Web*, 2017.

[25] Pei, K., Cao, Y., Yang, J. and Jana, S. "Deepxplore: Automated whitebox testing of deep learning systems," in *Proceedings of the 26th Symposium on Operating Systems Principles*, 2017, pp. 1–18.

[26] Samek, W., Wiegand, T., and Müller, K. R. "Explainable Artificial Intelligence: Understanding, Visualizing and Interpreting Deep Learning Models," *arXiv preprint arXiv:1708.08296*, 2017.

[27] Ćalušić, A. L. et al., "Biomarkers of mild hyperthermia related to flashover training in firefighters," *Journal of thermal biology*, vol. 37, no. 8, pp. 548–555, 2012.

[28] Ljubičić, A., Varnai, V. M., Petrinec, B., and Macan, J. "Response to thermal and physical strain during flashover training in Croatian firefighters," *Applied ergonomics*, vol. 45, no. 3, pp. 544–549, 2014.

[29] Gorbett, G. E., Hopkins, R., and Kennedy, P. "The current knowledge & training regarding backdraft, flashover, and other rapid fire progression phenomena," in *annual meeting of the National Fire Protection Association, Boston, MA*, 2007.

[30] Ebersole, J. F., Furlong, T. J., and Ebersole Jr, J. F. "Augmented reality-based firefighter training system and method," Dec-2002.

[31] Buckman III, J. M. *Chief fire officer's desk reference*. Jones & Bartlett Learning, 2005.

[32] Lee, S., and Harada, K. "A Simplified Formula for Occurrence of Flashover and Corresponding Heat Release Rate," *Procedia Engineering*, vol. 62, pp. 292–300, Jan. 2013.

[33] Johansson, N. and Hees, P. "A correlation for predicting smoke layer temperature in a room adjacent to a room involved in a pre-flashover fire," *Fire and Materials*, vol. 38, no. 2, pp. 182–193, 2014.

[34] Andersson, L. "Thermal Exposure Caused by the Smoke Gas Layer in Pre-flashover Fires: A Two-zone Model Approach," 2016.

[35] Sung, K. H., Choi, M. S., Hong, S. H., Park, K. W., and Ryou, H. S., "Numerical Study on the Prediction Method of Flashover in a Compartment," *Journal of Korean Society of Hazard Mitigation*, vol. 15, no. 4, pp. 123–128, 2015.

[36] Evegren, F., and Wickström, U. "New approach to estimate temperatures in pre-flashover fires: Lumped heat case," *Fire safety journal*, vol. 72, pp. 77–86, 2015.

[37] Kim, J. H. and Lattimer, B. Y. "Real-time probabilistic classification of fire and smoke using thermal imagery for intelligent firefighting robot," *Fire Safety Journal*, vol. 72, pp. 40–49, 2015.

[38] Kolomay, R. and Hoff, R. *Firefighter rescue & survival*. PennWell Books, 2003.

[39] Amon, F. and Pearson, C. "Thermal imaging in firefighting and thermography applications," *Radiometric Temperature Measurements: II. Applications*, vol. 43, pp. 279–331, 2009.

[40] Quiñones, M. A., Ford, J. K., and Teachout, M. S. "The relationship between work experience and job performance: A conceptual and meta-analytic review," *Personnel Psychology*, vol. 48, no. 4, pp. 887–910, 1995.

[41] Liao, H., Arvey, R. D., Butler, R. J., and Nutting, S. M. "Correlates of work injury frequency and duration among firefighters.," *Journal of occupational health psychology*, vol. 6, no. 3, p. 229, 2001.

[42] Mondragon-Gilmore, J. *Firefighters and the experience of increased intuitive awareness during emergency incidents*. Pacifica Graduate Institute, 2014.

[43] Tzeng, F. Y., and Ma, K. L. "Opening the black box-data driven visualization of neural networks," in *Visualization, 2005. VIS 05. IEEE*, 2005, pp. 383–390.

[44] Radford, A., Metz, L., and Chintala, S. "Unsupervised representation learning with deep convolutional generative adversarial networks," *arXiv preprint arXiv:1511.06434*, 2015.

[45] Sosnowski, T., Madura, H., Bieszczad, G., and Kastek, M. "Thermal camera system integrated into firefighter helmet," *Measurement Automation Monitoring*, vol. 63, 2017.

[46] Wu, R., Yan, S., Shan, Y., Dang, Q., and Sun, G. "Deep image: Scaling up image recognition," *arXiv preprint arXiv:1501.02876*, 2015.

[47] Rajpura, P., Goyal, M., Hegde, R., and Bojinov, H. "Dataset Augmentation with Synthetic Images Improves Semantic Segmentation," *arXiv preprint arXiv:1709.00849*, 2017.

[48] Taylor, L., and Nitschke, G. "Improving Deep Learning using Generic Data Augmentation," *arXiv preprint arXiv:1708.06020*, 2017.

[49] Xu, Y. *et al.*, "Improved relation classification by deep recurrent neural networks with data augmentation," *arXiv preprint arXiv:1601.03651*, 2016.

[50] Sun, C., Shrivastava, A., Singh, S., and Gupta, A. "Revisiting unreasonable effectiveness of data in deep learning era," in *2017*

*IEEE International Conference on Computer Vision (ICCV)*, 2017, pp. 843–852.

[51] Noh, Z., Sunar, M. S., and Pan, Z. "A review on augmented reality for virtual heritage system," in *International Conference on Technologies for E-Learning and Digital Entertainment*, 2009, pp. 50–61.

[52] Zhang, J., Li, Z., and Yang, J. "A divisional incremental training algorithm of support vector machine," in *Mechatronics and Automation, 2005 IEEE International Conference*, 2005, vol. 2, pp. 853–856.

[53] Goto, Y., Hochberg, M. M., Mashao, D. J., and Silverman, H. F., "Incremental MAP estimation of HMMs for efficient training and improved performance," in *Acoustics, Speech, and Signal Processing, 1995. ICASSP-95., 1995 International Conference on*, 1995, vol. 1, pp. 457–460.

[54] Shilton, A., Palaniswami, M., Ralph, D., and Tsoi, A. C. "Incremental training of support vector machines," *IEEE transactions on neural networks*, vol. 16, no. 1, pp. 114–131, 2005.

[55] Xiao, R., Wang, J., and Zhang, F. "An approach to incremental SVM learning algorithm," in *Tools with Artificial Intelligence, 2000. ICTAI 2000. Proceedings. 12th IEEE International Conference on*, 2000, pp. 268–273.